\documentclass[letterpaper]{article} 
\usepackage{amsfonts}
\usepackage{amsmath}
\usepackage{xcolor}
\usepackage{aaai25}  
\usepackage{times}  
\usepackage{helvet}  
\usepackage{courier}  
\usepackage[hyphens]{url}  
\usepackage{graphicx} 
\usepackage{natbib}  
\usepackage{caption} 
\usepackage{algorithm}
\usepackage{algorithmic}
\usepackage{booktabs}    
\usepackage{multirow}   
\usepackage{graphicx}
\usepackage{amsmath}
\usepackage{amssymb}
\usepackage{booktabs}
\usepackage[accsupp]{axessibility}  

\usepackage{tikz}
\usepackage{comment}
\usepackage{color}
\usepackage{graphicx}
\usepackage{multirow}
\usepackage{tabularx}
\usepackage{bbding}
\usepackage{float}
\usepackage{cite}
\usepackage{xspace}
\usepackage[misc]{ifsym}
\usepackage{bbding}
\usepackage{pifont}
\usepackage{colortbl}
\usepackage{wasysym}
\newcommand{\cmark}{\ding{51}\xspace}%
\newcommand{\xmarkg}{\textcolor{lightgray}{\ding{55}}\xspace}%
\defcitealias{D2Zero}{He et al. 2023}
\newcommand{\ours}{\textbf{ZoRI}\xspace}
\def\eg{\emph{e.g.}}

\usepackage{newfloat}
\usepackage{listings}
\DeclareCaptionStyle{ruled}{labelfont=normalfont,labelsep=colon,strut=off} 
\floatstyle{ruled}
\newfloat{listing}{tb}{lst}{}
\floatname{listing}{Listing}
\title{ZoRI: Towards Discriminative Zero-Shot Remote Sensing Instance Segmentation}
\author{
    Shiqi Huang\textsuperscript{\rm 1}\equalcontrib,
    Shuting He\textsuperscript{\rm 2}\equalcontrib,
    Bihan Wen\textsuperscript{\rm 1}\thanks{Corresponding author}
}
\affiliations{
    \textsuperscript{\rm 1}Nanyang Technological University\\
    \textsuperscript{\rm 2}Shanghai University of Finance and Economics\\

    shiqi006@e.ntu.edu.sg, heshuting555@gmail.com, bihan.wen@ntu.edu.sg
}

\begin{document}

\maketitle

\begin{abstract}
Instance segmentation algorithms in remote sensing are typically based on conventional methods, limiting their application to seen scenarios and closed-set predictions. In this work, we propose a novel task called zero-shot remote sensing instance segmentation, aimed at identifying aerial objects that are absent from training data. Challenges arise when classifying aerial categories with high inter-class similarity and intra-class variance. Besides, the domain gap between vision-language models' pretraining datasets and remote sensing datasets hinders the zero-shot capabilities of the pretrained model when it is directly applied to remote sensing images. To address these challenges, we propose a \textbf{Z}ero-Sh\textbf{o}t \textbf{R}emote Sensing \textbf{I}nstance Segmentation framework, dubbed \ours. Our approach features a discrimination-enhanced classifier that uses refined textual embeddings to increase the awareness of class disparities. Instead of direct fine-tuning, we propose a knowledge-maintained adaptation strategy that decouples semantic-related information to preserve the pretrained vision-language alignment while adjusting features to capture remote sensing domain-specific visual cues. Additionally, we introduce a prior-injected prediction with cache bank of aerial visual prototypes to supplement the semantic richness of text embeddings and seamlessly integrate aerial representations, adapting to the remote sensing domain. We establish new experimental protocols and benchmarks, and extensive experiments convincingly demonstrate that \ours achieves the state-of-art performance on the zero-shot remote sensing instance segmentation task. Our code is available at \url{https://github.com/HuangShiqi128/ZoRI}.

\end{abstract}

%
\section{Introduction}

Instance segmentation for remote sensing images aims to identify aerial objects and locate them with pixel-level masks for each instance \cite{su2019object, su2020hq,  carvalho2020instance, zhang2021semantic, xu2021improved, liu2024learning}. The task is of great importance for precise interpretation and comprehension of remote sensing images, which are essential for advanced earth observation. It has extensive real-world applications, such as military operations, marine monitoring, and urban planning \cite{article, de2021instance, wei2022lfgnet, yasir2023instance}, making it a pivotal research area in earth vision studies. 

\begin{figure}[t]
    \centering
	\vspace{-3mm}
        \includegraphics[width=0.46\textwidth]{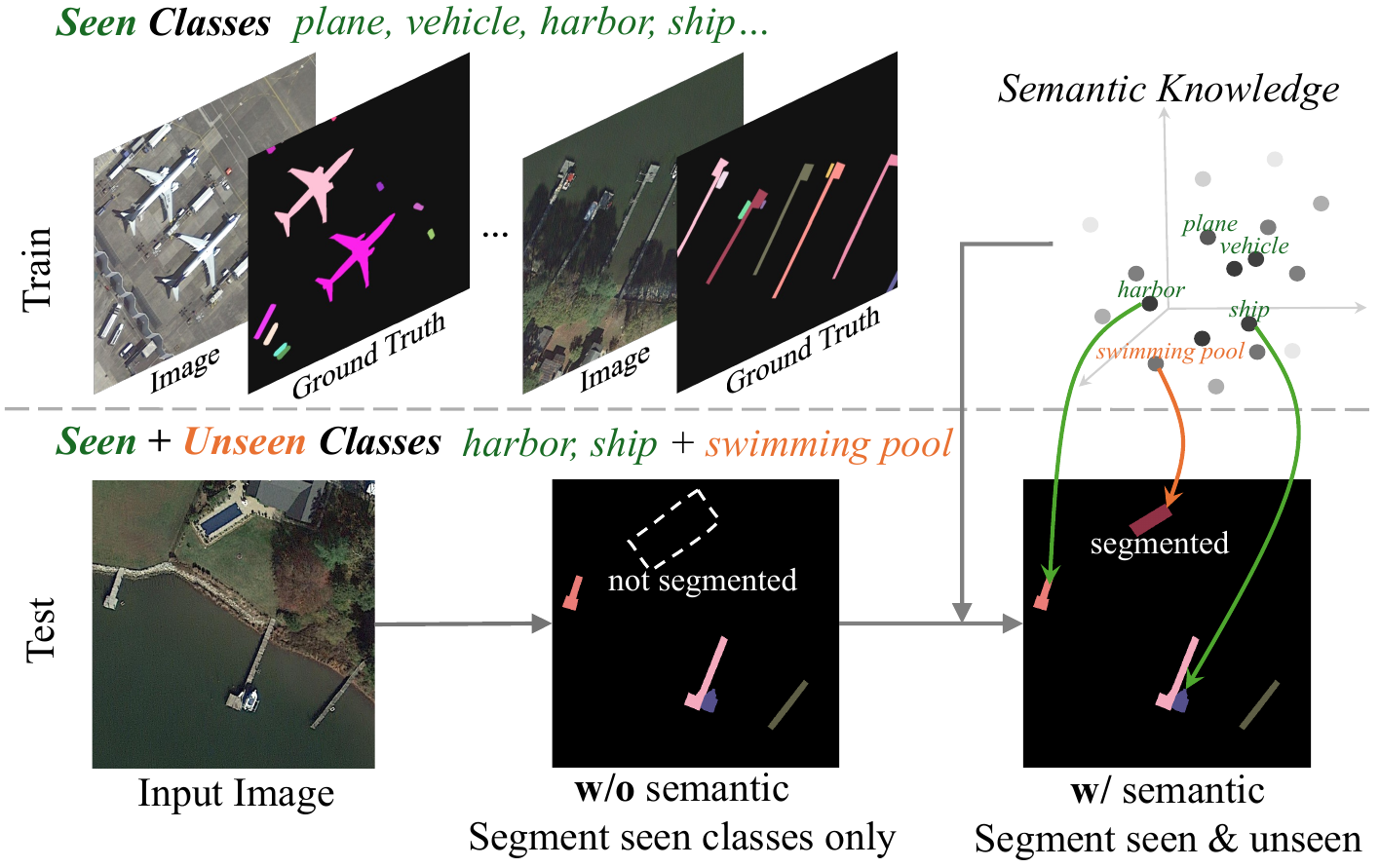}
	\vspace{-2mm}	
	\caption{Illustration of zero-shot remote sensing instance segmentation, which transfers the learned semantic knowledge from seen classes, \eg, \textit{harbor} and \textit{ship}, to the unseen class, \eg, \textit{swimming pool}.}
	\label{fig:teaser1}
	\vspace{-2mm}
\end{figure}

\begin{figure}[t]
    \centering
	\vspace{-3mm}
        \includegraphics[width=0.46\textwidth]{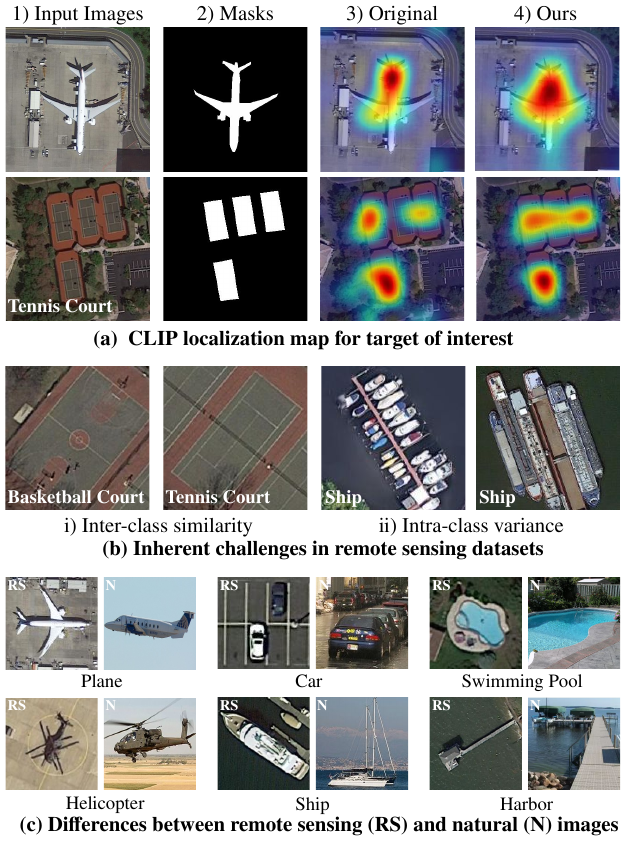}
	\vspace{-2mm}	
	\caption{\textbf{(a)} After refinement, the head of \textit{plane} is highlighted and the activation map strictly follows its shape; For \textit{tennis court}, the activation map is more focused and the missed one in the middle using original channels is also emphasized. \textbf{(b)} Classes such as \textit{basketball court} and \textit{tennis court} share similar color and shape, whereas instances from \textit{ship} can have various appearances. \textbf{(c)} Remote sensing images are in bird's eye view while natural images are from ground-level prospective.} 
	\label{fig:teaser2}
	\vspace{-6mm}
\end{figure}

The standard pipeline for aerial instance segmentation starts with annotating instance masks for remote sensing datasets, followed by training a segmentation model using conventional methods \cite{zamir2019isaid, su2019object}. However, due to relatively small object sizes and the large number of instances present in aerial images, labeling large scale datasets becomes time-consuming and cost-intensive. Consequently, applications are often constrained to limited scenarios. Moreover, training with conventional methods restricts the model to close-set predictions, making it unable to identify novel classes absent from training data.

We, for the first time, propose the zero-shot instance segmentation scheme for remote sensing images, which enables models to segment novel aerial object classes without instance mask annotations, as shown in Figure \ref{fig:teaser1}. The proposed scheme utilizes Vision-Language Models (VLMs) such as CLIP \cite{radford2021learning} which demonstrated remarkable capabilities in zero-shot tasks due to their highly aligned vision and language modalities \cite{ding2022decouplingzeroshotsemanticsegmentation, xu2022simplebaselineopenvocabularysemantic}. However, this alignment does not always extend consistently along channel dimensions \cite{zhu2023features}.
Our investigation using aerial images reveals that not all CLIP-extracted text features attend to the corresponding visual targets across these channels. As shown in Figure \ref{fig:teaser2}(a) 3), the activation region deviates from the object of interest when using the original CLIP text features. This indicates the pretrained CLIP struggles to establish a semantic-visual correspondence for remote sensing images. We further contend that this problem can be worsened by the inherent challenges of high inter-class (between-class) similarity and intra-class (within-class) variance \cite{zhang2020well, rsiscmdl, 9024005, rs15184588, yang2022scrdet++} in aerial objects, present in Figure \ref{fig:teaser2}(b). To address this, we propose a Discrimination-Enhanced Classifier (DEC), which refines the feature channels in the CLIP text embedding classifier to select most attentive ones to the target, encapsulating the most discriminative semantic information. After refinement, as shown in Figure \ref{fig:teaser2}(a) 4), the activation map accurately depicts the shape of the \textit{plane} and highlights the middle \textit{tennis court} missed in Figure \ref{fig:teaser2}(a) 3). 

While directly applying pretrained VLMs achieves success with natural image, it fails to produce similar gains due to the inherent domain gap between VLM pretraining and remote sensing datasets. Specifically, as shown in Figure \ref{fig:teaser2}(c), aerial images shot from above have different appearances compared to natural images, which are typically captured from a ground-level prospective and have a central layout. To bridge this gap, we design a Knowledge-Maintained Adaptation (KMA) with proposed decoupling strategy for CLIP visual features to adapt CLIP to the remote sensing domain without drastically forgetting pretrained knowledge or losing generalization ability. After decoupling, cross-modal alignment is preserved by retaining the decoupled semantic-related feature channels, while non-semantic-related knowledge is learned to adapt to the specific visual domain of remote sensing. This design not only maintains the well-trained discrimination power of CLIP but also derives tailored visual representations for aerial images. 

Furthermore, although our predictions become more discriminative using DEC, text embeddings in the classifier still represent general concepts for natural images and convey limited information, leading to reduced classification capability with current zero-shot classifier. To inject aerial-specific knowledge, we introduce a Prior-Injected Prediction (PIP), which utilizes visual prototypes to complement the final prediction. By incorporating a cache bank constructed with remote sensing visual prototypes, domain-specific visual information is seamlessly integrated into the model. Text meanings in the original classifier are re-weighted to align with remote sensing characteristic, and visual information is added to supplement various class representations that cannot be delivered by text alone. 

Our main contributions are summarized as follows:
\begin{itemize}
    \item We introduce the problem of zero-shot remote sensing instance segmentation. To overcome inherent challenges in aerial imagery and the domain gap, we develop a \textbf{Z}ero-sh\textbf{o}t \textbf{R}emote sensing \textbf{I}nstance segmentation framework (\ours) and establish new experimental protocols and benchmarks. \ours achieves the state-of-art performance.
    \item To address the classification ambiguity for aerial objects, we propose DEC, which refines CLIP textual embeddings to enhance the discriminative power, improving the ability to classify different classes more distinctly.
    \item We design KMA with a decoupled visual feature learning strategy to adapt CLIP to the remote sensing domain while preserving its vision-language alignment.
    \item We identify the limitation of classifying with text embeddings and introduce PIP using a cache bank consisting of aerial visual prototypes to enhance semantic richness, thereby narrowing the representation gap between general textual expressions and remote sensing objects.
\end{itemize}

\section{Related Work}
\subsection{Instance Segmentation in Remote Sensing}
Segmentation frameworks for remote sensing images have seen significant advancements. Various methods have been developed to address challenges such as severe scale variations, foreground-background confusion, and class ambiguity \cite{liu2024learning, ye2023remote, su2022faster, su2020hq, su2019object, 9204465, zhang2021semantic} in remote sensing instance segmentation datasets. Recently, the integration of vision foundation models with remote sensing image segmentation has also been evolving. For instance, RSPrompter \cite{chen2024rsprompter} improves the instance segmentation performance of SAM \cite{kirillov2023segany} on remote sensing images through prompt learning. Additionally, SAMRS \cite{SAMRS} expands remote sensing segmentation datasets by leveraging SAM. However, current algorithms for remote sensing image segmentation primarily focus on close-set prediction, limiting their ability to recognize a broader array of object categories that are not present in the training set.

\subsection{Zero-shot Learning in Remote Sensing}
Zero-shot learning has attracted significant research interest in the field of remote sensing. Earlier works primarily untilized language models like Word2vec \cite{mikolov2013efficientestimationwordrepresentations} and BERT \cite{devlin2019bertpretrainingdeepbidirectional} to extract semantic representations for zero-shot scene classification \cite{li2017zero, li2021learning, wang2021distance}. Recently, with the emergence of vision-language models such as CLIP \cite{radford2021learning}, research has shifted towards constructing highly-aligned vision-language embedding space \cite{li2023rs, remoteclip, zavras2024mindmodalitygapremote, mall2023remotesensingvisionlanguagefoundation}. DescReg \cite{zang2024zeroshotaerialobjectdetection} presents a zero-shot aerial object detection framework by introducing visual descriptions to regularize semantic embeddings retrieved from BERT. However, zero-shot remote sensing instance segmentation remains unexplored.

\subsection{Zero-shot Segmentation}
Zero-shot segmentation aims to identify objects beyond seen classes during training with pixel-wise masks. Owing to the powerful generalization capability of CLIP, many zero-shot segmentation models are developed by harnessing its zero-shot classification ability. Previous methods \cite{ding2022decouplingzeroshotsemanticsegmentation, xu2022simplebaselineopenvocabularysemantic, liang2023open, SegPoint} retrieve class-agnostic masks with a mask generator and classify using a CLIP-based classifier. To avoid duplicate feature extractions for mask generation and CLIP classification, ZegCLIP \cite{zhou2022zegclip} extends CLIP zero-shot capability from image-level to pixel-level by deep prompt tuning. FC-CLIP \cite{yu2023convolutions} further demonstrates the efficacy of zero-shot classification using the frozen convolutional CLIP backbone. Zero-shot instance segmentation was first introduced in ZSI \cite{zheng2021zeroshotinstancesegmentation}, followed by works \cite{PADing, D2Zero} to handle classification bias and ambiguity issues.

\section{Method}
\begin{figure*}[t]   
	\centering
	\includegraphics[width=\linewidth,scale=1.00]{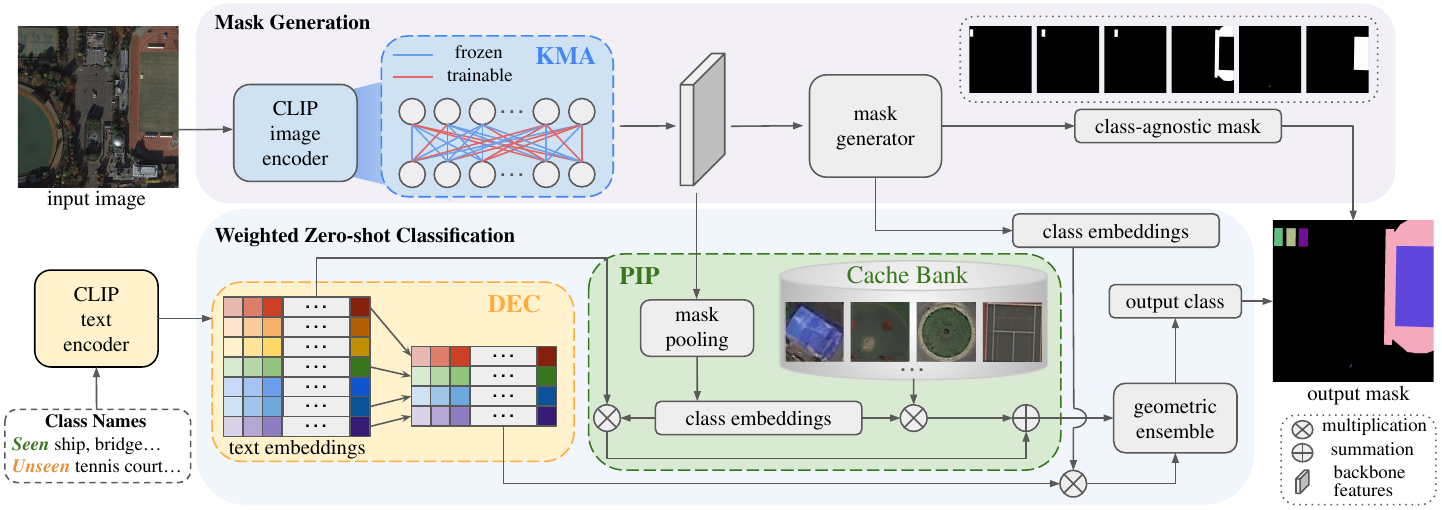}
	\caption{Overview of \ours framework. The CLIP image encoder is partially trained with knowledge-maintained adaptation (KMA) to extract backbone features, which are then fed into a mask generator to produce mask predictions and class embeddings. Discriminative-enhanced classifier (DEC) constructed by refining the original text embeddings is then used to classify class embeddings. During inference, cache bank is incorporated into CLIP zero-shot predictions to derive prior-injected predictions (PIP). The final classification probability is obtained through an ensemble approach.}
	\label{fig:framework}
 \vspace{-15pt}
\end{figure*}

\subsection{Problem Definition}
In zero-shot remote sensing instance segmentation (ZSRI), there are $N$ classes divided into two groups, $N^s$ seen classes denoted as $C^s$ and $N^u$ unseen classes denoted as $C^u$, with $C^s \cap C^u = \text{\O}$ and $N^s+N^u=N$, where $C^s = \{c^s_1,c^s_2,..., c^s_{N^s}\}$, $C^u = \{c^u_1,c^u_2,..., c^u_{N^u}\}$. The training set comprises images $x_s$ with seen classes $C^s$ but no object of unseen classes $C^u$, along with pixel-wise instance mask annotations for each object in $C^s$. During inference, two settings are considered: ZSRI and Generalized ZSRI (GZSRI). In ZSRI setting, the goal is to segment objects of only $N^u$ unseen classes, while in GZSRI, the object is to segment all $N^u + N^s$ classes, including both seen and unseen, which is more reflective of real-world scenarios.

\subsection{Architecture Overview}
The architecture of our proposed \ours is illustrated in Figure \ref{fig:framework}. We adopt the paradigm of FC-CLIP \cite{yu2023convolutions}, developed based on Mask2Former \cite{cheng2022maskedattention}. The CLIP image encoder is partially trained with knowledge-maintained adaptation (KMA) to extract backbone features, which are then fed into a mask generator to produce mask predictions and class embeddings. Text embeddings are encoded by the frozen CLIP text encoder and further refined to construct discriminative-enhanced classifier (DEC), which is used to classify class embeddings. During inference, cache bank is incorporated into CLIP zero-shot predictions to derive prior-injected predictions (PIP). The final classification probability is obtained through an ensemble approach similar in FC-CLIP, by combining scores based on trained class embeddings, and prior-injected predictions based on mask-pooled CLIP backbone features.

\subsection{Discrimination-Enhanced Classifier (DEC)}
Considering the inherent challenges present in remote sensing classes, such as high inter-class similarity and intra-class variance, a naive classifier constructed from text embeddings struggles to overcome the resulting classification ambiguity. 
To address this, we propose a discrimination-enhanced classifier, which refines the naive classifier by selecting the most discriminative channels, discarding redundant ones which are less attentive to object targets and may introduce noise during classification. The resulting text embeddings in the classifier will have an improved one-to-one correspondence with the target objects.

For each category $c_i$ from $C^s \cup C^u$, we obtain its category text embedding by placing the category name into prompt templates, \eg, `satellite imagery of \{\}.',  `aerial imagery of \{\}.', and inputting to the CLIP text encoder. The final text embedding for each class is obtained by averaging the CLIP-extracted features across all templates, denoted as $x_i \in \mathbb{R}^D$, where $i \in \{1, 2,..., N\}$. The zero-shot classifier is then constructed from text embeddings $[x_1\ x_2\ ...\ x_N] \in \mathbb{R}^{D\times N}$. 

To refine the classifier, we select top-k channels from $D$ dimensions that minimize similarity and maximize variance across different classes. The similarity among channels across classes is represented as:
\begin{align}
    S = \frac{1}{N\times(N-1)}\sum^N_{i=1}\sum^N_{j=1,j\neq i}x_i x_j \in \mathbb{R}^D,
\end{align}
where $x_i$ and $x_j$ are L2-normalized.
The variance is formulated as:
\begin{align}
    V = \frac{1}{N}\sum_{i=1}^N (x_i - \bar x)^2 \in \mathbb{R}^D,
\end{align}
where $\bar x = \frac{1}{N}\sum_{i=1}^N x_i$ is the average across categories.
Taking both similarity and variance into account, we formulate our objective as:
\begin{align}
    O = -\lambda S + (1-\lambda)V \in \mathbb{R}^D,
\label{eq:objective}
\end{align}
where $\lambda$ is a balance factor.
The indices corresponding to the top-k largest values in $O \in \mathbb{R}^D$ are selected as the refined feature channels. These channels possess the highest discirminative power by suppressing resemblance and enhancing distinctiveness among different categories, thereby are most attentive to object targets.
 
\subsection{Knowledge-Maintained Adaptation (KMA)}
The shared frozen convolutional CLIP backbone \cite{yu2023convolutions} offers a favorable balance between accuracy and computational cost by combining feature extraction for mask generation and CLIP zero-shot prediction. However, this frozen pretrained CLIP vision backbone struggles with generating good representations for aerial images. VLMs like CLIP are pretrained on large-scale web-based image-text pairs predominantly featuring natural images captured at ground level, leading to a domain gap when applied to aerial images. Objects viewed from a bird's-eye view exhibit different appearances compare to those in natural images, which are typically captured from a ground-level prospective and have a central layout. Applying the frozen CLIP backbone directly inevitably overlooks such difference and fails to use training priors to improve feature extraction for aerial images. To adapt the pretrained backbone to remote sensing domain, directly fine-tuning tends to overfit to seen classes, resulting in drastically forgetting pretrained knowledge and compromising the model's ability to handle new, unseen classes \cite{zhou2022maskclip, xu2023side, zhou2022zegclip}. To transfer to remote sensing domain without compromising generalization ability, we propose a decoupling strategy that separates visual features into two groups. The first group includes channels more closely related to semantics, which exhibit better vision-language alignment and are thus more discriminative in classifying objects. The second group consists of channels less connected to semantic meaning but rich in visual cues.

Object instances for each seen class from $C^s$ can be cropped from the downstream training dataset with ground truth annotations. Suppose we retrieve $T$ instances for each class, and extract intermediate backbone features for these instances with the pretrained CLIP vision encoder. To decouple backbone feature channels into two groups, we follows the criterion present in Eq.(\ref{eq:objective}). Similarity and variance of backbone features among categories of $T$ instances are calculated in a manner similar to text embeddings $x_i$. Channels with lowest similarity and highest variance are considered as the first group. Due to its highly aligned vision and language embedding space, we freeze them to keep them intact which helps maintain the pretrained cross-modality alignment. The remaining channels are less related to semantics and have knowledge in visual representation, thus we make them trainable to adapt to remote sensing specific visual domain. After decoupling, visual representations can be tuned to adapt to remote sensing domain while preserving cross-modal alignment. This design not only maintains the well-trained discrimination power of CLIP, but also derives tailored visual representation for aerial images. 

\subsection{Prior-Injected Prediction (PIP)}
Current classification results solely rely on text embeddings extracted from category descriptions using the CLIP text encoder. However, due to the high intra-class variability within remote sensing images, objects in the same category can have diverse appearances. Text features convey limited information and are insufficient to encompass this variability. Consequently, text classifier may struggle to establish a reliable correspondence between various aerial visual features and pretrained textual features, resulting in less discriminative ability. Therefore, we propose to introduce visual prototypes to enhance classification capability by covering various vision samples through a cache bank. By integrating the cache bank constructed with visual prototypes, the classification score derived from text classifier is re-weighted to align with remote sensing domain. This approach incorporates aerial-specific visual prior to enhance the representation of various classes, addressing the limitations that cannot be delivered through text alone. 

\subsubsection{Cache Bank Construction}
To construct a cache bank with supplementary visual representations specific to remote sensing domain, we first crop object instances for each seen class $c^s_i \in \{c^s_1,c^s_2,..., c^s_{N^s}\}$ from the downstream training dataset with ground truth annotations. Suppose we retrieve $K$ sample instances for each class, denoted as $\{I_n\}_{n=1}^{KN^s}$, these instances are regarded as representatives for different categories in $\{c^s_1,c^s_2,..., c^s_{N^s}\}$ in aerial images. Visual embeddings for sample instances $\{I_n\}_{n=1}^{KN^s}$ are then obtained using the pretrained CLIP image encoder $E_{vis}$. Then, we concatenate the training-set instance visual features along the sample dimension, and store them as the cache bank in remote sensing domain, formulated as:
\begin{align}
    F = \texttt{concat}(\{E_{vis}(I_n)\}_{n=1}^{KN^s}) \in \mathbb{R} ^{KN^s\times D},
\end{align}
where $D$ denotes the feature dimension of the vision encoder.
The cache bank $F$ is constructed in a training-free manner similar to works in few-shot learning \cite{rong2023retrieval, zhang2021tip, zhang2023prompt}. Such bank comprises diverse representations retrieved from remote sensing visual examples, which augment CLIP with aerial knowledge, different from natural images for its bird's eye view.

For $N^u$ unseen classes whose ground truth annotations are assumed unavailable, we use $P$ predictions with the highest probability as visual samples for unseen categories. Then, visual embeddings are also retrieved using the pretrained CLIP image encoder and concatenated to the cache bank, which is updated to obtain $F \in \mathbb{R}^{(KN^s+PN^u)\times D}$.

\subsubsection{Cache Bank Prediction}
After getting mask predictions, mask-pooling is applied to the CLIP-extracted backbone feature for the input image, denoted as $f \in \mathbb{R}^{1\times D}$ for each predicted segment. The mask-pooled CLIP backbone feature serves as the query, and the pre-encoded instance features $F$ can be regarded as keys. The cosine similarity metric between the query and keys is calculated as: 
\begin{align}
    M = f F^T \in \mathbb{R}^{1\times (KN^s+PN^u)},
\end{align}
where mask-pooled feature and cache bank features are L2-normalized. Along with the one-hot class label $l \in \mathbb{R}^{1\times N}$ associated with each sample instance in the cache bank, values for query and key are thus constructed by concatenating one-hot labels, denoted as $L \in \mathbb{R}^{(KN^s+PN^u)\times N}$. We thus formulate the cache bank prediction as below:
\begin{align}
    \text{logits}_{\text{cb}} = \texttt{softmax}(M)L \in \mathbb{R}^{1\times N}.
\end{align}
The cache bank prediction is fused into CLIP zero-shot prediction to derive the prior-injected prediction with the formulation:
\begin{align}
    \text{logits}_{\text{pip}} = fW^T + \alpha \text{logits}_{\text{cb}} \in \mathbb{R}^{1\times N},
\end{align}
where $W \in \mathbb{R}^{N\times D}$ is CLIP's classifier, $fW^T$ calculates the classification score based on pretrained zero-shot prediction from CLIP, $\alpha$ is a balance factor.

\section{Experiments}

\subsection{Experimental Setup}
\subsubsection{Datasets}
We establish two remote sensing zero-shot instance segmentation benchmarks with iSAID \cite{zamir2019isaid} and NWPU-VHR-10 \cite{cheng2014multi,su2019object} datasets. iSAID is a large-scale dataset for instance segmentation in aerial images containing 15 classes. NWPU-VHR-10 is a very high resolution (VHR) object detection dataset \cite{cheng2014multi} further annotated with instance masks \cite{su2019object}, which contains 10 classes. iSAID dataset is divided into 11 seen classes and 4 unseen classes (‘tennis court’, ‘helicopter’, ‘swimming pool’ and ‘soccer ball field'), which has the same seen/unseen split for DOTA \cite{zang2024zeroshotaerialobjectdetection, Xia_2018_CVPR}, and NWPU-VHR-10 dataset is split into 7 seen classes and 3 unseen classes (‘ship’, ‘basketball court’ and `harbor’). For the training set, only images containing seen class objects are selected, while any images with unseen classes are excluded to avoid information leakage. Both ZSRI and GZSRI are evaluated for testing. Dataset details and more datasets can be found in the supplementary material.

\subsubsection{Evaluation Metrics}
Referring to previous works \cite{zheng2021zeroshotinstancesegmentation, D2Zero}, we use Recall@100, i.e., top 100 instances, across different IoU thresholds \{0.4, 0.5, 0.6\} and mean Average Precision (mAP) with IoU threshold 0.5 as the metrics. For GZSRI setting, harmonic mean (HM) \cite{ xian2020zeroshotlearningcomprehensive} of seen and unseen classes is computed to indicate overall performance.

\subsubsection{Implementation Details}
The proposed method is developed based on FC-CLIP \cite{yu2023convolutions}. All hyperparameters remain unchanged unless otherwise specified. We use the LAION-2B pretrained ConvNext-Large \cite{liu2022convnet2020s} from OpenCLIP \cite{ilharco_gabriel_2021_5143773} as the feature extractor. The mask generator follows Mask2Former \cite{cheng2022maskedattention} with object query number set to 300. Prompt templates for RESISC45 \cite{cheng2017remote} used in CLIP \cite{radford2021learning} are employed to obtain text embeddings with the pretrained CLIP text encoder. We train the model for 50 epochs with training batch size 2. Input images are resized to $512\times 512$ during training. Hyper-parameters $\lambda$ and $\alpha$ are set to 0.7, 0.5, respectively. Instance number $T$ and trainable channels in KMA is empirically set to 1 and 32. The model is optimized using AdamW optimizer. The learning rate is set to $1.25\times 10^{-5}$. All experiments are conducted with one RTXA5000 GPU.

\subsection{Component Analysis}
We conduct extensive experiments to demonstrate the effectiveness of proposed modules in Table \ref{tab:ablation_study} with both iSAID \cite{zamir2019isaid} and NWPU-VHR-10 \cite{cheng2014multi} datasets. FC-CLIP \cite{yu2023convolutions} is used as our baseline with replaced prompt templates RESISC45 for remote sensing images to derive text embeddings for the classifier. From the baseline result, we can see the performance on unseen classes lags behind seen classes by a large margin, which indicates the model does not generalize well on aerial images. In the following sections, we provide both quantitative and qualitative analysis of our proposed modules. By integrating all the modules, we achieve a final result that significantly outperforms the baseline.

\begin{table}[t]
\setlength\tabcolsep{3.5pt}
\scriptsize
  \begin{center}
  \begin{tabular}{cccccccccc}
  \toprule
  \multirow{2}[6]{*}{Dataset} & \multirow{2}[6]{*}{DEC} & \multirow{2}[6]{*}{KMA} & \multirow{2}[6]{*}{PIP} & \multicolumn{3}{c}{mAP $\uparrow$} & \multicolumn{3}{c}{Recall@100 $\uparrow$} \\
   \cmidrule(r){5-7}     \cmidrule(r){8-10}
  &  &  & &  Seen  & Unseen & HM & Seen  & Unseen & HM \\
  \toprule
\multirow{4}{*}{iSAID} 
& \xmarkg& \xmarkg& \xmarkg&  43.47  &  4.91  &  8.83  &  67.01  &  36.31  &  47.10\\
& \cmark& \xmarkg& \xmarkg &  46.56  &  8.28  &  14.05  &   68.24   &  40.24  & 50.62 \\
& \cmark& \cmark& \xmarkg &  45.90  &  8.64  &  14.55   &  68.14  &  \textbf{41.52}  &  \textbf{51.60} \\
& \cmark& \cmark& \cmark &  \textbf{47.05}  &  \textbf{9.30}  &  \textbf{15.53}   &  \textbf{68.89}  &  37.73  &  48.76\\
  \arrayrulecolor{gray}\hline
\multirow{4}{*}{NWPU} 
& \xmarkg & \xmarkg & \xmarkg& 80.85 &  6.86  &  12.65 &   94.25 &  36.16  &  52.27 \\
& \cmark& \xmarkg& \xmarkg &  \textbf{82.90}  &  8.81  &  15.93  &  \textbf{94.81}  &  38.11  &  54.37 \\
& \cmark& \cmark& \xmarkg &  80.65 &  9.32 &  16.70  &   93.53   &  \textbf{46.44}   &  \textbf{62.07} \\
& \cmark& \cmark& \cmark &  80.57  & \textbf{12.26}  & \textbf{21.28} & 93.80 & 44.01 & 59.91 \\
 \arrayrulecolor{black}\bottomrule
  \end{tabular}
\vspace{-3mm}
\caption{Component Analysis of our \ours under GZSRI setting. DEC represents discrimination-enhanced classifier. KMA is knowledge-maintained adaptation. PIP denotes prior-injected prediction.}
\label{tab:ablation_study}
\vspace{-6mm}
\end{center}
\end{table}

\begin{table}[t]
\setlength\tabcolsep{4.5pt}
\scriptsize
  \begin{center}
  \begin{tabular}{c|cccccccc}
  \toprule
  \multicolumn{2}{c}{\#Channel} & 200 & 300 & 400  & 500 &  600  & 700 & 768\\
  \toprule
  \multirow{3}{*}{mAP $\uparrow$} 
  & Seen & 45.61 & \textbf{46.56} & 44.99 & 45.09 & 44.39 & 43.69 & 43.47\\
  & Unseen & 6.26 & \textbf{8.28} & 8.08 & 7.09 & 6.59 & 6.19 & 4.93\\
  & HM & 11.02 & \textbf{14.06} & 13.69 & 12.26 & 11.47 & 10.86 & 8.86\\
  \bottomrule
  \end{tabular}
\vspace{-3mm}
\caption{Ablation study on number of selected channels in text classifier on iSAID dataset.}
\label{tab:channel}
\vspace{-7mm}
\end{center}
\end{table}

\subsubsection{Discrimination-Enhanced Classifier}
By selecting the most discriminative channels in CLIP text classifier, we observe enhancements in both seen and unseen classes in Table \ref{tab:ablation_study}. Due to these universal improvements, there is a notable increase of 5.22\% HM-mAP for iSAID and 3.28\% HM-mAP for NWPU-VHR-10. The improvements in both seen and unseen classes indicate that the classifier can effectively capture the most deterministic feature for each aerial object. As a result, the weak semantic correspondence and classification ambiguity caused by high inter-class similarity and intra-class variance in remote sensing categories is reduced, allowing for more accurate recognition of class instances. 

In Table \ref{tab:channel}, we conduct an ablation study to evaluate the impact of the number of selected channels on the text classifier's performance on iSAID dataset. The original text embeddings have 768 channels. The results indicate that there is an optimal range where the classifier's discriminatory power reaches its peak. Beyond this range, performance declines due to excessive information loss or insufficient disentanglement of the text embeddings. From the result, we thereby select the top-300 channels with the highest score formulated in Eq.(\ref{eq:objective}).

\subsubsection{Knowledge-Maintained Adaptation}
By integrating visual feature decoupling for knowledge-maintained adaptation with DEC, the mAP for unseen classes sees a cumulative improvement of 3.73\% on iSAID dataset. The HM-recall increases by nearly 10\% for NWPU dataset. The improvement demonstrates that the CLIP vision encoder has successfully adapted to the remote sensing domain and can generate tailored representations for aerial objects. It is worth noting that the unseen mAP shows a significant improvement for both datasets, indicating the well-pretrained vision-language alignment is preserved during the adaptation process, maintaining the model's zero-shot ability to handle novel classes.

To validate our adaptation method, we report the results using other finetuning techniques in Table \ref{tab:kma}. Compared with full finetuning, we achieve a larger seen mAP while only train a few parameters and our unseen mAP is about 1.5\% higher. This indicates our method is more effective in adaptation while preserving pretrained knowledge. We also report the result using VPT \cite{jia2022vpt} following their implementation for ConvNext. From the results, we can see that KMA outperforms VPT in both seen and unseen classes, with a 7.33\% increase in HM-recall. Our advantage demonstrates the efficacy of our proposed module KMA.

\begin{table}[t]
\setlength\tabcolsep{6pt}
\scriptsize
  \begin{center}
  \begin{tabular}{lcccccc}
  \toprule
  \multirow{2}[6]{*}{Method} & \multicolumn{3}{c}{mAP $\uparrow$} & \multicolumn{3}{c}{Recall@100 $\uparrow$} \\
   \cmidrule(r){2-4}     \cmidrule(r){5-7}
  & Seen  & Unseen & HM & Seen & Unseen & HM \\
  \toprule
baseline &  43.47  &  4.91  &  8.83  &  67.01  &  36.31  &  47.10 \\
baseline+FT & 43.97 &  5.71  & 10.10  &  \textbf{67.33}  &  36.69  & 47.50 \\
baseline+VPT &  40.93  &  7.18  &  12.22  &  62.46  &  33.46  &  43.57  \\
baseline+KMA&  \textbf{44.08} &  \textbf{7.19} &  \textbf{12.36}  &  67.19  &  \textbf{40.96}  &  \textbf{50.90} \\
 \arrayrulecolor{black}\bottomrule
  \end{tabular}
\vspace{-3mm}
\caption{Comparison with other finetuning methods on iSAID dataset. The baseline method uses a frozen vision encoder, while baseline+FT denotes full finetuning of the vision encoder.}
\label{tab:kma}
\vspace{-6mm}
\end{center}
\end{table}

\subsubsection{Prior-Injected Prediction}
Incorporating a cache bank of visual priors into the final class prediction leads to performance improvements on both iSAID and NWPU datasets, as seen from Table \ref{tab:ablation_study}. Specifically, on the iSAID dataset, there is an overall 6.7\% gain in HM-mAP when combined with DEC and KMA compared to the baseline. Similarly, on the NWPU dataset, there is a 8.63\% improvement in HM-mAP. This enhancement demonstrates that integrating visual samples through the cache bank boosts the model's accuracy by exposing it to a wider variety of remote sensing objects. The visual prototypes effectively address the limitations of text descriptions by considering the intra-class variability of aerial objects, thus facilitating better adaptation to the remote sensing domain.

\begin{figure*}[ht]
    \centering
	\includegraphics[width=0.96\textwidth]{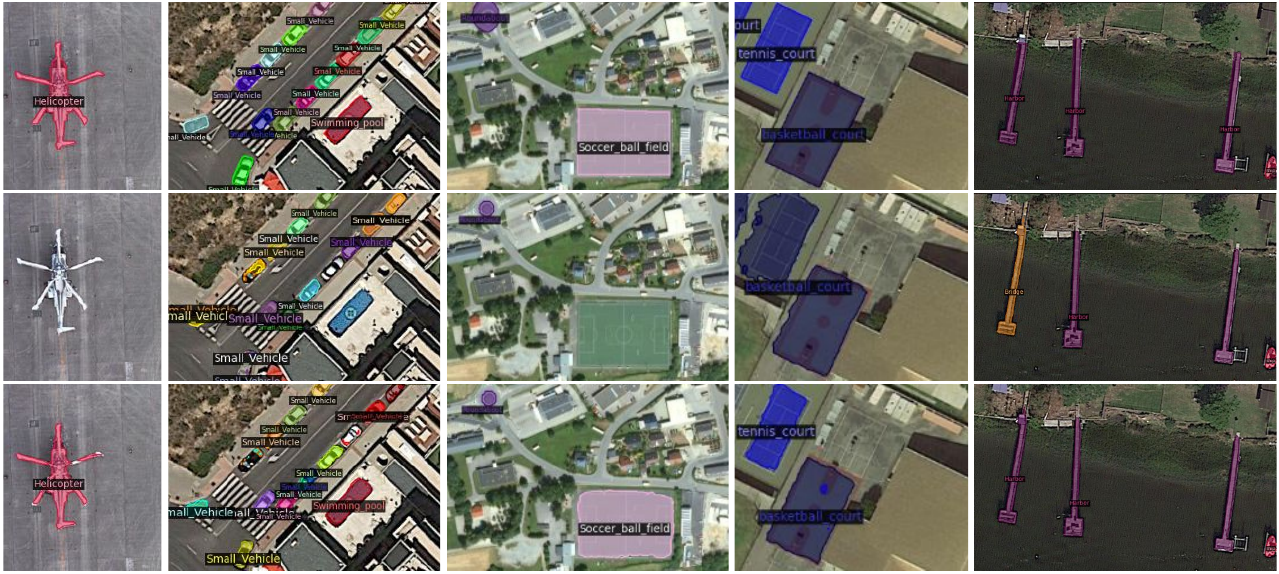}
	\vspace{-3mm}
	\caption{Comparison of GZSRI results: (top row) ground truth, (middle row) FC-CLIP~\cite{yu2023convolutions} and (bottom row) our results. \ours successfully segments unseen objects missed by FC-CLIP due to domain gap, \eg, \textit{helicopter}, \textit{swimming pool}, and \textit{soccer ball field} in the first three columns, and correctly identifies similar categories misclassified by FC-CLIP, \eg, \textit{tennis court} and \textit{harbor} in the last two columns due to class ambiguity. The proposed method \ours shows much better results by constructing a more discriminative model adapting to the remote sensing domain.} 
	\vspace{-6mm}
	\label{fig:demo}
\end{figure*}

To examine how cache size influences prediction accuracy, we create cache bank with varying numbers of visual prototypes, and report the results in Table \ref{tab:cache}. The results indicate that a moderate number of prototypes from each class is enough to improve the performance, and only 1 sample cannot capture sufficient variability. To balance computational cost and accuracy, we choose 4 visual prototypes.

\vspace{-2mm}
\begin{table}[t]
\setlength\tabcolsep{6pt}
\scriptsize
  \begin{center}
  \begin{tabular}{c|cccccccc}
  \toprule
  \multicolumn{2}{c}{Cache Size} & 0  & 1 & 4 & 8  & 16 &  32\\
  \toprule
  \multirow{3}{*}{mAP $\uparrow$} 
  & Seen & 45.90 & 46.07 & 47.05 & 47.09 & 47.09 & \textbf{47.11}\\
  & Unseen & 8.64 & 8.69 & \textbf{9.30} & 9.27 & 9.29 & 9.28\\
  & HM & 14.55 & 14.62 & \textbf{15.53} & 15.49 & 15.52 & 15.51\\
  \bottomrule
  \end{tabular}
\vspace{-3mm}
\caption{Ablation study on cache size on iSAID dataset.}
\label{tab:cache}
\vspace{-7mm}
\end{center}
\end{table}

\begin{table}[t]
\setlength\tabcolsep{2pt}
\scriptsize
  \begin{center}
  \begin{tabular}{clcccccc}
  \toprule
  \multirow{2}[6]{*}{Dataset} & \multirow{2}[6]{*}{Method} & \multicolumn{3}{c}{mAP $\uparrow$} & \multicolumn{3}{c}{Recall@100 $\uparrow$} \\
   \cmidrule(r){3-5}     \cmidrule(r){6-8}
  & &  Seen  & Unseen & HM & Seen & Unseen & HM \\
  \toprule
\multirow{5}{*}{\cellcolor{white}iSAID} 
& ZSI \cite{zheng2021zeroshotinstancesegmentation} &   45.83 &  0.64   &  1.26  &   61.13  &   16.66   & 26.19 \\
& D$^2$Zero \citepalias{D2Zero} & 39.80  &  0.39  &  0.77   &  63.45  & 10.16   & 17.52 \\
& FC-CLIP \cite{yu2023convolutions} &  43.47  &  4.91  &  8.83  &  67.01  &  36.31  &  47.10 \\
& \ours&  \textbf{47.05}  &  \textbf{9.30}  &  \textbf{15.53}   &  \textbf{68.89}  &  \textbf{37.73}  &  \textbf{48.76}\\
& \textcolor{gray}{Supervised} &   \textcolor{gray}{52.40} &  \textcolor{gray}{0.0}  & \textcolor{gray}{-}    &   \textcolor{gray}{72.89}  & \textcolor{gray}{0.0}    & \textcolor{gray}{-} \\
\arrayrulecolor{gray}\hline
\multirow{5}{*}{\cellcolor{white}NWPU}
& ZSI \cite{zheng2021zeroshotinstancesegmentation}&  \textbf{83.01} &   0.13  &  0.25   &  89.90   &   11.10  & 19.76  \\
& D$^2$Zero \citepalias{D2Zero}&   77.62 &  0.41  & 0.81   & \textbf{95.41}    &  9.78   & 17.75 \\
& FC-CLIP \cite{yu2023convolutions}&  80.85 &  6.86  &  12.65 &   94.25 &  36.16  &  52.27\\
& \ours&  80.57  & \textbf{12.26}  & \textbf{21.28} & 93.80 & \textbf{44.01} & \textbf{59.91} \\
& \textcolor{gray}{Supervised} &   \textcolor{gray}{83.87} &  \textcolor{gray}{0.0}  & \textcolor{gray}{-}    &   \textcolor{gray}{95.58}  & \textcolor{gray}{0.0}    & \textcolor{gray}{-} \\
 \arrayrulecolor{black}\bottomrule
  \end{tabular}
\vspace{-3mm}
\caption{Comparison under GZSRI setting.}
\label{tab:sota_gzsri}
\vspace{-7mm}
\end{center}
\end{table}

\subsection{Comparison with State-of-the-Art Methods}
In Table \ref{tab:sota_gzsri} and Table \ref{tab:sota_zsri}, we compare with SOTA methods on iSAID and NWPU-VHR-10 datasets, evaluated under the GZSRI and ZSRI settings respectively. We also provide the supervised results by replacing CLIP text classifier to the conventional trainable classifier for reference. As we can see from the tables, our proposed \ours surpasses all compared methods. Under the GZSRI setting, our approach achieve a notable improvement of 14.27\% HM-mAP over ZSI \cite{zheng2021zeroshotinstancesegmentation} on the iSAID dataset and 21.03\% HM-mAP on the NWPU-VHR-10 dataset. Additionally, we outperform D$^2$Zero by 14.76\% and 20.47\% HM-mAP on each dataset. Compared with FC-CLIP, we achieve 6.7\% and 8.63\% gain on HM-mAP on iSAID and NWPU respectively. Under the ZSRI setting, \ours obtains more than 5\% gain in mAP and significant increases in recall across various thresholds compared to the best-reported method on both datasets. 

Figure \ref{fig:demo} shows a qualitative comparison with FC-CLIP on the iSAID dataset for both seen and unseen classes. Our method successfully segments the unseen objects missed out by FC-CLIP, such as \textit{helicopter}, \textit{swimming pool}, and \textit{soccer ball field} in the first three columns. This is because our model adapts to remote sensing domain and has better knowledge for aerial objects. Notably, we correctly identify similar categories like \textit{tennis court} and \textit{basketball court}, \textit{harbor} and \textit{bridge}, which are misclassified by FC-CLIP. This highlights the effectiveness of our discrimination-enhanced classifier. The proposed method \ours shows much better results by constructing a more discriminative model adapting to remote sensing domain. 

\begin{table}[t]
\setlength\tabcolsep{7pt}
\scriptsize
  \begin{center}
  \begin{tabular}{clcccc}
\toprule
 \multirow{2}[6]{*}{Dataset} & \multirow{2}[6]{*}{Method} & mAP $\uparrow$ &\multicolumn{3}{c}{Recall@100 $\uparrow$}\\
 \cmidrule(r){4-6}
 &  &  0.5  & 0.4 & 0.5 & 0.6 \\
  \toprule
   \multirow{4}{*}{iSAID}
   & ZSI \cite{zheng2021zeroshotinstancesegmentation} &  1.5 & 22.6 & 16.7 & 10.6\\
   & D$^2$Zero  \citepalias{D2Zero}&{0.5} &{36.6} &{28.1} &{19.4}\\
   & FC-CLIP \cite{yu2023convolutions}&{3.0} &{51.4} &{42.2} &{30.8}\\
   &\ours & \textbf{8.6} & \textbf{55.2} & \textbf{46.0} & \textbf{36.6} \\
  \arrayrulecolor{gray}\hline
  \multirow{4}{*}{NWPU}
   & ZSI \cite{zheng2021zeroshotinstancesegmentation}&  0.6 & 13.9 & 11.1 & 7.9\\
   & D$^2$Zero  \citepalias{D2Zero}&{2.7} &{23.9} &{19.2} &{14.3}\\
   & FC-CLIP \cite{yu2023convolutions}&{4.5} &{46.2} &{37.1} &{30.8}\\
   & \ours &\textbf{9.6} &\textbf{55.9} &\textbf{47.3} &\textbf{33.8}\\
  \arrayrulecolor{black}\bottomrule
  \end{tabular}
\vspace{-3mm}
\caption{Comparison under ZSRI setting.}
\label{tab:sota_zsri}
\vspace{-8mm}
\end{center}
\end{table}

\section{Conclusion}
We propose a new zero-shot remote sensing instance segmentation task and present a novel framework \ours. Discrimination-enhanced classifier is developed to tackle the inherent challenges posed by high inter-class similarity and intra-class variance in remote sensing classes. To adapt CLIP to remote sensing domain without losing its generalization power, we introduce knowledge-maintained adaptation based on a visual feature decoupling strategy to derive tailored representations for aerial objects. Additionally, our prior-injected prediction, which integrates a cache bank of visual prototypes, supports text embedding classifier with diverse aerial objects. With new benchmarks established, \ours achieves the state-of-the-art performance in zero-shot remote sensing instance segmentation task. This advancement not only pushes the boundaries of zero-shot learning in remote sensing but also sets the stage for future research in zero-shot segmentation of aerial imagery.

\section{Acknowledgments}
This work was supported in part by the Economic Development Board, Republic of Singapore, through its Space Technology Development Programme Grant under Grant S22-02004-STDP.

\bibliographystyle{aaai25}

\clearpage
\section{Supplementary Material for ZoRI}
\section{A. Dataset Details}
The details of the three remote sensing datasets including one additional dataset are as follows:
\begin{itemize}
    \item \textbf{iSAID} \cite{zamir2019isaid} is a large-scale aerial instance segmentation dataset containing 655,451 object instances for 15 classes across 2,806 high-resolution images. It has same classes as DOTA-v1.0 \cite{Xia_2018_CVPR}. The spatial resolution of images ranges from 800 to 13, 000 in width.
    \item \textbf{NWPU-VHR-10} \cite{cheng2014multi} is a 10-class geospatial object detection dataset. It contains a total of 800 VHR optical remote sensing images divided into two sets: a) positive image set contains 650 images with at least one target in each image, b) negative image set contains 150 images and it does not contain any targets. The instance mask annotations are further added and provided by Su et al \cite{su2019object}.
    \item \textbf{SIOR} \cite{SAMRS} is a large scale remote sensing segmentation dataset provided in SAMRS \cite{SAMRS}. It is developed from DIOR \cite{Li_2020} with instance mask annotations generated from SAM \cite{kirillov2023segany}. It has 23463 images with size 800 $\times$ 800 for 20 classes. We do not report its results in the main paper since it is generated with pseudo mask annotations from SAM, which does not have ground truth masks as iSAID and NWPU-VHR-10 datasets. However, we believe it is still valuable to include these results in the supplementary material for reference.
\end{itemize}

\subsection{Train/test Split}
For iSAID dataset, we use the original training and validation set for our training and testing, as they did not publicly release test set annotations. There are 1411 images for training and 458 images for testing. For NWPU-VHR-10 dataset, as they did not split the dataset, we random split the images in the positive image set with train-test ratio 0.8. After the split for 650 positive images, we have 520 images in the training set and 130 images in the testing set. For SIOR dataset, we use the \texttt{txt} files provided in SAMRS to get train and validation splits. There are 11725 training images and 11738 testing images.

\subsection{Image Preprocessing}
For iSAID and NWPU-VHR-10, due to their varying image size, we crop images in the datasets into smaller patches use the dataset preparation codes provided in iSAID develop kit. The patch size is set to $800 \times 800$ for iSAID dataset following the official usage with a stride set to 200. Due to the relatively small dataset size for NWPU-VHR-10 dataset, we set the patch size to $512 \time 512$ to enlarge the training dataset. With the cropping, we obtain 18732/9512 and 2654/731 images (train/test) for iSAID and NWPU-VHR-10 respectively. No image preprocessing is applied to SIOR since images are already in shape 800 $\times$ 800.

\subsection{Seen/unseen Split} 
For iSAID dataset, as it has same classes as DOTA-v1.0, we follow DescReg \cite{zang2024zeroshotaerialobjectdetection} to split the 15 classes into 11/4 for seen/unseen classes. For NWPU-VHR-10 dataset, to ensure an meaningful seen/unseen split, we perform hierarchical clustering on the class semantic embeddings and sample one class for pairs of leaf nodes in the clustering tree. This is to maintain diversity in unseen classes and evenly distribute semantically similar classes between seen and unseen sets, challenging the model to effectively predict both seen and unseen classes. The resulting split is 7/3 for seen/unseen classes. For SIOR dataset, as it is developed from DIOR and thus has same classes as DIOR, we follow RRFS \cite{huang2022robustregionfeaturesynthesizer} to split the 20 classes in to 16/4 for seen/unseen classes. Classes after seen/unseen split are:
\begin{itemize}
    \item \textbf{iSAID dataset} \\
    \textbf{Seen}: `ship', `storage tank', `baseball diamond', `basketball court', `ground track field', `bridge',
                `large vehicle', `small vehicle', `roundabout', `plane', `harbor'. \\
    \textbf{Unseen}: `tennis court', `helicopter', `swimming pool', `soccer ball field'.
    \item \textbf{NWPU-VHR-10 dataset} \\
    \textbf{Seen}: `airplane', `storage tank', `baseball diamond', `tennis court', `ground track field', `bridge', `vehicle'.
    \textbf{Unseen}: `ship', `basketball court', `harbor'.
    \item \textbf{SIOR dataset} \\
    \textbf{Seen}: `airplane’, `baseball field’, `bridge’, `chimney’, `dam’, `expressway service area’, `expressway toll station’, `golf field’, `harbor’, `overpass’, `ship’, `stadium’, `storage tank’, `tennis court’, `train station’, `vehicle’. \\
    \textbf{Unseen}: `airport’, `basketball court’, `ground track field’, `windmill’.
    
\end{itemize}

\subsection{Annotation Preparation}
For zero-shot remote sensing instance segmentation, we prepare and create annotation files tailored for zero-shot training setting. For the training annotation file, all annotations for unseen classes are removed and images containing any unseen classes are discarded from the file. Only seen classes data are included in the file. The filtered annotation files for training are saved as \textit{isaid\_seen\_11\_4\_train.json}, \textit{nwpu\_seen\_7\_3\_train.json}, and \textit{sior\_seen\_16\_4\_train.json} for iSAID, NWPU-VHR-10, and SIOR, respectively.
For testing under GZSRI setting, the original test annotations, including both seen and unseen classes, are used for evaluation. In contrast, for the ZSRI setting, only annotations for unseen classes are retained. Resulting testing annotation files are saved as \textit{isaid\_gzsri\_val.json}, \textit{isaid\_unseen\_11\_4\_val.json} for iSAID dataset, \textit{nwpu\_gzsri\_val.json} and \textit{nwpu\_unseen\_7\_3\_val.json} for NWPU-VHR-10 dataset, \textit{sior\_gzsri\_val.json}, \textit{sior\_unseen\_16\_4\_val.json} for SIOR dataset . After filtering, there are 15524/9512, 2053/731 and 8729/11738 images (train/test) with usable annotations for iSAID, NWPU-VHR-10, and SIOR, respectively.

\section{B. Implementation Details}
\subsection{\ours Implementation} 
\subsubsection{Training} 
 We use the LAION-2B pretrained ConvNext-Large \cite{liu2022convnet2020s} from OpenCLIP \cite{ilharco_gabriel_2021_5143773} as the feature extractor. The intermediate backbone features have 192 channel dimensions, among which we freeze 160 channels and train the rest 32 channels to adapt to remote sensing domain. Empirically, we found that there is no significant difference when varying the number of instances from each class to determine channels. The mask generator follows Mask2Former \cite{cheng2022maskedattention} with object query number set to 300. Prompt templates for RESISC45 \cite{cheng2017remote} used in CLIP \cite{radford2021learning} are employed to obtain text embeddings with the pretrained CLIP text encoder. Discrimination-enhanced classifier, which contains the refined text embeddings, is utilized for the training. We choose the top-300 channels which minimize the similarity and maximize the variance across text embeddings for different classes based on the ablation study present in the paper. During training, class embedding from mask generator is also trained to align with the 300 selected channels from text embeddings. We train the model for 50 epochs with training batch size 2. Input images are resized to $512\times 512$ during training. Hyper-parameter $\lambda$ is set to 0.7. The model is optimized using AdamW optimizer with weight decay 0.05. The learning rate is set to $1.25\times 10^{-5}$ with multi-step decay schedule. All experiments are conducted with one RTXA5000 GPU.

\subsubsection{Testing}
The Prior-Injected Prediction using cache bank is calculated during inference. For unseen classes, we use predictions with the highest probability after incorporating DEC and KMA to get visual samples. Due to the presence of numerous inaccurate predictions, which could introduce incorrect samples into the cache bank, we only use the top-1 probability prediction as the pseudo visual sample for unseen classes. To balance cache samples for seen and unseen classes when computing cache bank logits, we repeat the top-1 unseen features to match half the number of cache size. We empirically found a moderate size for unseen class yields the best mAP and recall performance. The balance factor $\alpha$ for cache bank prediction is set to 0.5. 

\subsection{SOTA Methods Implementation}
We follow the implementation procedure stated in ZSI \cite{zheng2021zeroshotinstancesegmentation}. We adopt Word2vec  \cite{mikolov2013efficientestimationwordrepresentations} to get semantic embeddings for each class in our datasets, and normalize with L2-normalization. Semantic embedding for category with more than one word, such as \textit{swimming pool}, takes the average of word vectors across each word. The mean of semantic embeddings for all classes is used as the background class. The model is trained on images resized to 512, and max number of detections is set to 300 for testing. The other training settings and hyper-parameters are identical to ZSI. For D2Zero \cite{D2Zero} and FC-CLIP \cite{yu2023convolutions}, we employ RESISC45 prompt templates used in CLIP \cite{radford2021learning} to retrieve CLIP text embeddings for the classifier. Images are cropped to 512 $\times$ 512 for training and the query number is set to 300. We train both models for 50 epochs with training batch size 2 on one RTXA5000 GPU. The learning rate is set to $1.25\times 10^{-5}$ with multi-step decay schedule. The other training settings and specific hyper-parameters for each model remain unchanged.

\begin{figure}[t]
    \centering
	\includegraphics[width=0.46\textwidth]{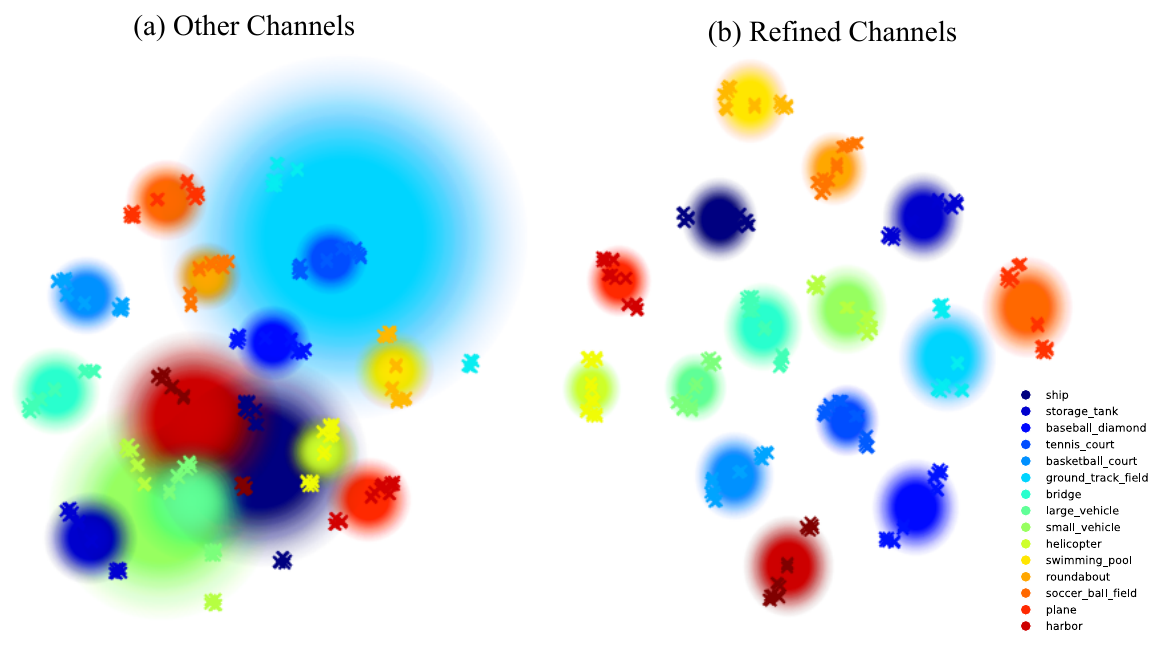}
	\caption{(Best viewed in color) t-SNE visualization of text embeddings. Crosses of the same color represent text embeddings produced using different prompt templates, with each class distinguished by a different color. The surrounding circle is added for better visualization, with the centroid representing the mean of text embeddings from the same class, and the radius being the maximum distance between any text embedding and the centroid.} 
 \label{fig:tsne}
\end{figure}

\section{C. Additional Results}
\subsection{Discrimination-Enhanced Classifier}
To compare our classifier with refined channels over others, we utilize t-SNE \cite{van2008visualizing} to visualize text embeddings with other channels and refined channels in Figure \ref{fig:tsne}. Each color represents the text embeddings for a specific category. Crosses of the same color represent text
embeddings produced using different prompt templates. From Figure \ref{fig:tsne}(a), we can see some text embeddings within the same class deviate from the main cluster, which can be considered as outliers. These outliers negatively impact the classification. To highlight this deviation, we enclose text embeddings for each class within a surrounding circle, where the centroid represents the mean of text embeddings for the class, and the radius being the maximum distance between any text embedding and the centroid. In Figure \ref{fig:tsne}(a), it is observed that some text embeddings are highly entangled since their circles are overlapped. For example, the embeddings for \textit{ship}, \textit{harbor}, \textit{small vehicle} and \textit{large vehicle} are overlapped, making their semantic differences barely noticeable. Similar pattern can also be observed for \textit{ground track field}, \textit{tennis court} and \textit{baseball diamond}. After refinement, as shown in Figure \ref{fig:tsne}(b), the text embeddings for different categories are pulled further away from each other and more distinctly separated, showing improved inter-class disentanglement. The distinctiveness of different categories enhances the model's robustness in identifying classes with high inter-class similarity and improves error-tolerance for instances with high intra-class variance.

\begin{table}[H]
\vspace{2mm}
\setlength\tabcolsep{3.5pt}
\scriptsize
  \begin{center}
  \begin{tabular}{cccccccccc}
  \toprule
  \multirow{2}[6]{*}{Dataset} & \multirow{2}[6]{*}{DEC} & \multirow{2}[6]{*}{KMA} & \multirow{2}[6]{*}{PIP} & \multicolumn{3}{c}{mAP $\uparrow$} & \multicolumn{3}{c}{Recall@100 $\uparrow$} \\
   \cmidrule(r){5-7}     \cmidrule(r){8-10}
  &  &  & &  Seen  & Unseen & HM & Seen  & Unseen & HM \\
  \toprule
\multirow{8}{*}{iSAID} 
& \xmarkg& \xmarkg& \xmarkg&  43.47  &  4.91  &  8.83  &  67.01  &  36.31  &  47.10\\
& \cmark& \xmarkg& \xmarkg &  46.56  &  8.28  &  14.05  &   68.24   &  40.24  & 50.62 \\
& \xmarkg& \cmark& \xmarkg &  44.08  &  7.19  &  12.36  &   67.19   &  \underline{40.96}  & \underline{50.89} \\
& \xmarkg& \xmarkg& \cmark &  44.44  &  5.16  &  9.24  &   67.80   &  32.85  & 44.26 \\
& \cmark& \cmark& \xmarkg &  45.90  &  8.64  &  14.55   &  68.14  &  \textbf{41.52}  &  \textbf{51.60} \\
& \cmark& \xmarkg& \cmark &  \textbf{47.64}  &  8.80  &  14.86   &  \textbf{68.90}  &  37.64  &  48.68 \\
& \xmarkg& \cmark& \cmark &  45.05  &  7.94  &  13.50   &  67.92  &  37.84  &  48.60 \\
& \cmark& \cmark& \cmark &  \underline{47.05}  &  \textbf{9.30}  &  \textbf{15.53}   &  \underline{68.89}  &  37.73  &  48.76\\
  \arrayrulecolor{gray}\hline
\multirow{8}{*}{NWPU} 
& \xmarkg & \xmarkg & \xmarkg& 80.85 &  6.86  &  12.65 &   94.25 &  36.16  &  52.27 \\
& \cmark& \xmarkg& \xmarkg &  \textbf{82.90}  &  8.81  &  15.93  &  \underline{94.81}  &  38.11  &  54.37 \\
& \xmarkg& \cmark& \xmarkg &  81.06  &  7.08  &  13.03  &  92.57  &  \underline{44.07}  &  59.71 \\
& \xmarkg& \xmarkg& \cmark &  80.83  &  8.49  &  15.37  &  94.72  &  32.84  &  48.77 \\
& \cmark& \cmark& \xmarkg &  80.65 &  9.32 &  16.70  &   93.53   &  \textbf{46.44}   &  \textbf{62.07} \\
& \cmark& \xmarkg& \cmark &  \underline{82.70} &  \underline{10.24} &  \underline{18.23}  &   \textbf{94.95}   &  35.81   &  52.00 \\
& \xmarkg& \cmark& \cmark &  81.03 &  8.18  &  14.85  &   93.27   &  40.99  &  56.95 \\
& \cmark& \cmark& \cmark &  80.57  & \textbf{12.26}  & \textbf{21.28} & 93.80 & 44.01 & \underline{59.91} \\
  \arrayrulecolor{gray}\hline
\multirow{8}{*}{SIOR} 
& \xmarkg & \xmarkg & \xmarkg& 50.35 &  6.55  &  11.60 &  75.22 &  \underline{45.80}  &  \underline{56.94} \\
& \cmark& \xmarkg& \xmarkg &  51.27  &  6.76  &  11.95  &  75.00  &  44.02  &  55.48 \\
& \xmarkg& \cmark& \xmarkg &  48.67  &  6.84  &  12.00  &  74.91  &  41.30  &  53.24 \\
& \xmarkg& \xmarkg& \cmark &  49.90  &  6.74  &  11.88 &  75.02  &  43.81 &  55.32 \\
& \cmark& \cmark& \xmarkg &  50.65 &  7.49 &  13.05  &   75.17   &  \textbf{46.01}   &  \textbf{57.08} \\
& \cmark& \xmarkg& \cmark &  \textbf{52.44} & \underline{7.77} &  \underline{13.54}  &   \underline{75.32}  &  42.27 & 54.15\\
& \xmarkg& \cmark& \cmark &  49.91 &  7.76  &  13.43  &   75.27   &  39.07  &  51.44 \\
& \cmark& \cmark& \cmark &  \underline{51.88}  & \textbf{8.48} & \textbf{14.57} & \textbf{75.58} & 44.17 & 55.76 \\
 \arrayrulecolor{black}\bottomrule
  \end{tabular}
\vspace{-3mm}
\caption{Component-wise results of our \ours under GZSRI setting. DEC represents discrimination-enhanced classifier. KMA is knowledge-maintained adaptation. PIP denotes prior-injected prediction. Results in \textbf{bold} are the best, results \underline{underlined} are the 2nd best.}
\label{tab:component}
\end{center}
\end{table}

\subsection{Component-wise Results}
We provide a more detailed component-wise results under GZSRI for iSAID, NWPU-VHR-10, and SIOR dataset in Table \ref{tab:component}. From the results, we can see that each component can improve the overall performance for mAP, validating the effectiveness of our proposed modules. The average precision consistently improves, but recall slightly decreases after integrating PIP. We believe this occurs because the introduced visual prototypes, while providing accurate visual samples for remote sensing objects, also tend to constrain the model, making it more likely to predict samples similar to those in the cache bank.
In all, \ours achieves much better overall performance on mAP across three benchmarks.

\subsection{Qualitative Results}
\begin{figure*}[htbp]
    \centering
	\includegraphics[width=0.96\textwidth]{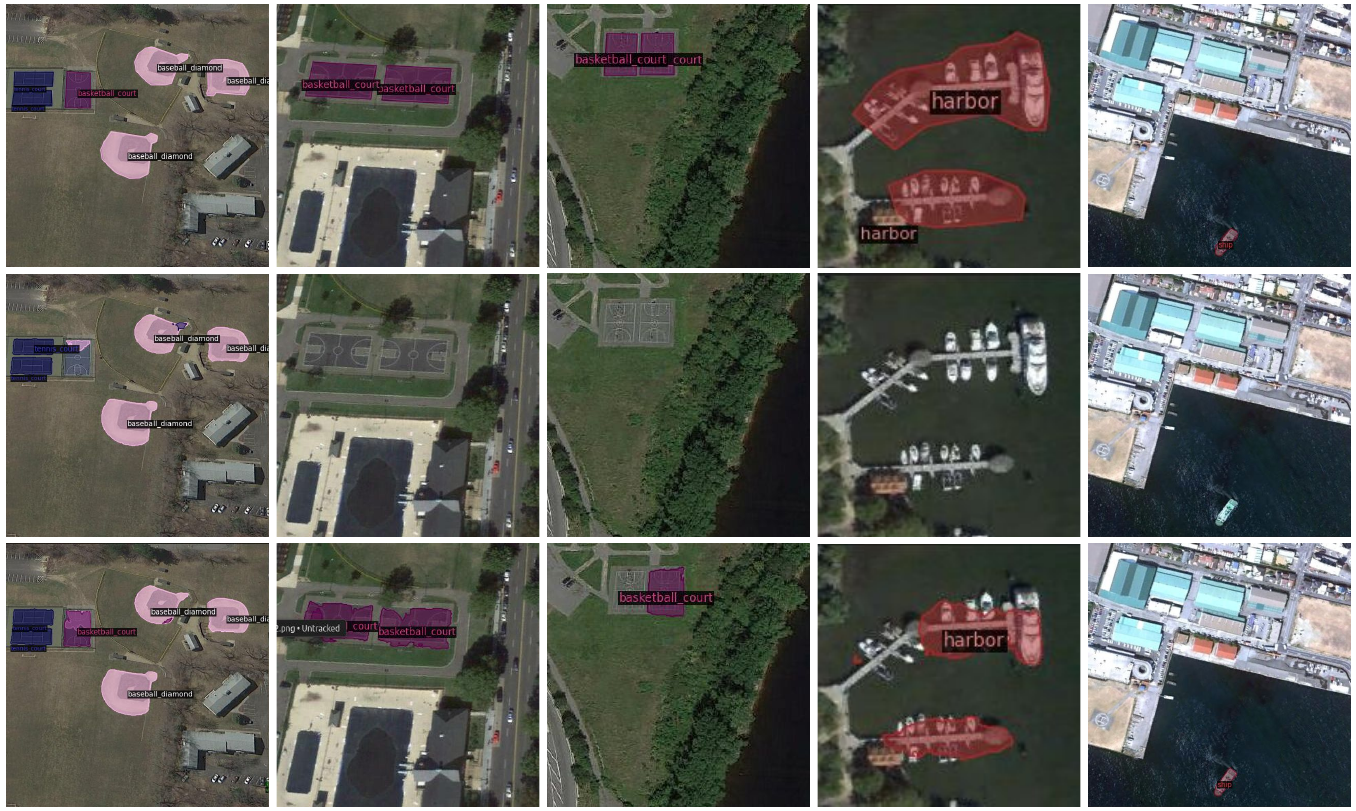}
        \vspace{-3mm}
	\caption{Comparison of GZSRI results on NWPU-VHR-10 dataset. \ours successfully segments unseen objects missed by FC-CLIP, \eg, \textit{basketball court} (columns 1-3), \textit{habor} (column 4), and \textit{soccer ball field} (last column).}
	\label{fig:demo1}
\end{figure*}

\begin{figure*}[htbp]
    \centering
	\includegraphics[width=0.96\textwidth]{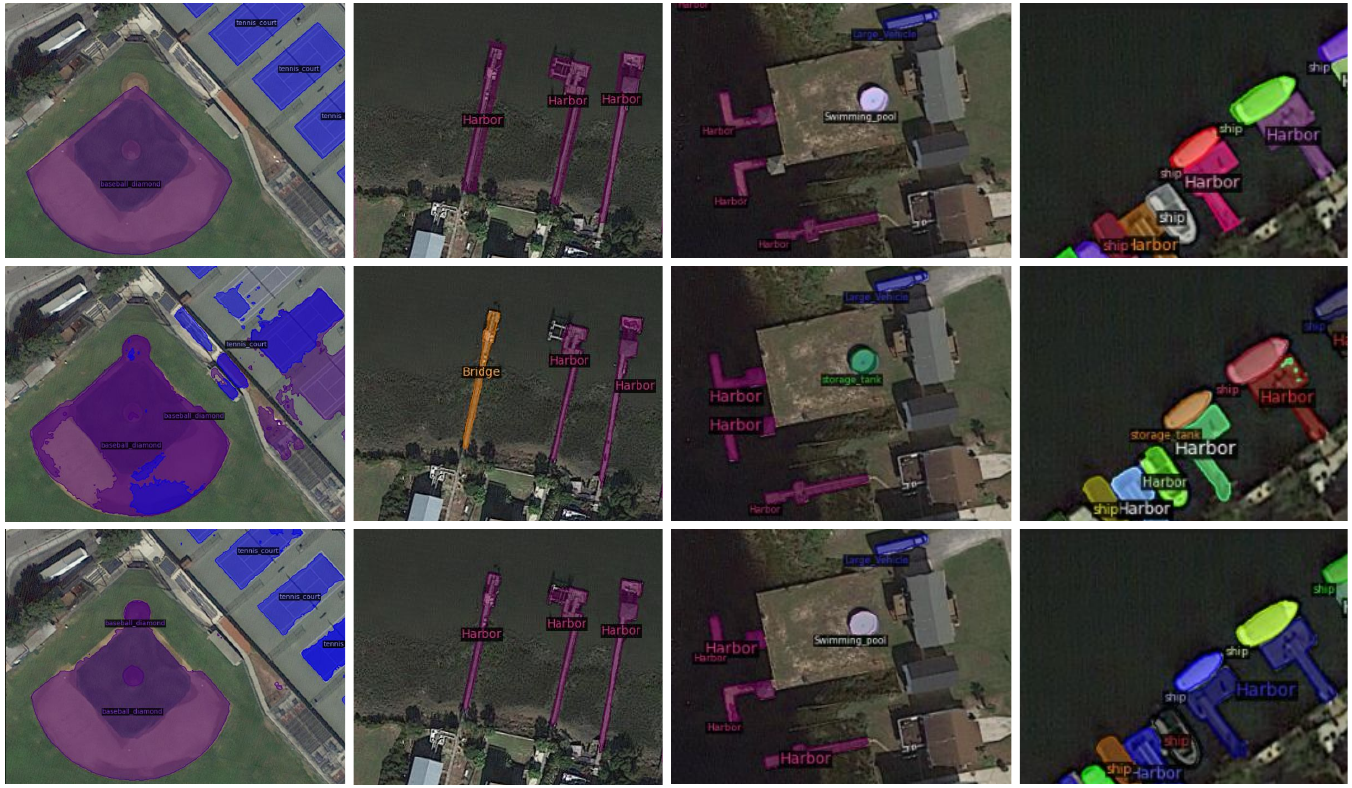}
        \vspace{-3mm}
	\caption{Comparison of GZSRI results on iSAID dataset. \ours achieves better segmentation results for unseen class \textit{tennis court} and seen class \textit{baseball diamond} (column 1), and correctly identifies similar categories misclassified by FC-CLIP, \eg, \textit{habor} and \textit{bridge} (column 2), \textit{swimming pool} and \textit{storage tank} (column 3), \textit{ship} and \textit{storage tank} (column 4).} 
	\label{fig:demo2}
\end{figure*}

Figure \ref{fig:demo1} shows a qualitative comparison with FC-CLIP on the NWPU-VHR-10 dataset. More qualitative results for iSAID dataset are showing in Figure \ref{fig:demo2}. From first to last row are: ground truth, FC-CLIP \cite{yu2023convolutions}, and our results. Kindly refer to figure captions for details.

\end{document}